\documentclass[11pt]{article}
\usepackage[a4paper,margin=1in]{geometry}
\usepackage{setspace}
\usepackage[T1]{fontenc}
\usepackage[utf8]{inputenc}
\usepackage{amsmath,amssymb,amsfonts}
\usepackage{bm}
\usepackage{graphicx}
\graphicspath{{figures/}}
\usepackage{booktabs}
\usepackage{array}
\usepackage{tabularx}
\usepackage{url}
\usepackage{cite}
\usepackage{xcolor}
\usepackage{hyperref}
\hypersetup{
  colorlinks=true, linkcolor=black, citecolor=black, urlcolor=blue!60!black,
  pdftitle={Skill Availability and Presentation Granularity in Large-Language-Model Agents: A Controlled SkillsBench Study},
  pdfauthor={Xiaonan Xu and Wenjing Wu},
  pdfsubject={Large-language-model agents, agent skills, SkillsBench},
  pdfkeywords={Agent skills, benchmark evaluation, large language models, reproducibility, SkillsBench, statistical inference}}
\usepackage{authblk}
\usepackage{titlesec}
\titleformat{\section}{\normalfont\large\bfseries}{\thesection.}{0.6em}{}
\titleformat{\subsection}{\normalfont\normalsize\bfseries}{\thesubsection.}{0.6em}{}
\usepackage[font=small,labelfont=bf,skip=4pt]{caption}
\title{\bfseries Skill Availability and Presentation Granularity in Large-Language-Model Agents: A Controlled SkillsBench Study}
\author[1]{Xiaonan Xu\thanks{Corresponding author: \texttt{xiaonanxu5@gmail.com}}}
\author[2]{Wenjing Wu}
\affil[1]{Computer Information Technology, Northern Arizona University, Flagstaff, AZ, USA}
\affil[2]{Computer Science, University of Colorado Boulder, Boulder, CO, USA}
\date{}

\begin{document}
\maketitle

\begin{abstract}
\noindent
Skill documents provide procedural knowledge to large-language-model agents at inference time. This article studies whether the presentation granularity of controlled skill knowledge changes downstream task success. The experiment uses a pinned SkillsBench version, a 30-task domain-balanced subset validated by official oracle runs, two reasoning-enabled model configurations, six skill conditions, and five trials per task-condition-model cell. Skill availability is the clearest empirical signal. Relative to no skill, skill conditions increase task-mean pass rate by 26.7 to 36.0 percentage points for GPT-5.5 and by 18.0 to 26.0 percentage points for DeepSeek V4-Flash. The final data contain 1,800 rows, with 900 rows for each model. The task is the inference unit. Five trials are aggregated within each task-condition-model cell before paired contrasts are estimated over 30 tasks. The primary presentation contrasts are smaller and uncertain. Low-abstraction guidance differs from high-abstraction guidance by +0.7 percentage points for GPT-5.5 and -6.7 percentage points for DeepSeek V4-Flash, with both 95\% bootstrap confidence intervals crossing zero. Adding one worked example to medium-abstraction guidance differs from the no-example variant by +0.7 and +1.3 percentage points. Mean-reward robustness checks preserve the same substantive conclusion. In this controlled subset, skill availability is associated with higher success than no skill, while the tested presentation-granularity changes yield small, uncertain, and model-dependent effects.

\vspace{0.8em}

\vspace{0.4em}
\noindent\textbf{Funding:} This work received no external funding.
\end{abstract}

\section{Introduction}
\label{sec:introduction}
Large language model (LLM) agents combine text-based reasoning with actions in external environments. The action space may include shell commands, code execution, file editing, browser automation, and deterministic task verifiers. ReAct-style prompting made the coupling between reasoning and action explicit, while tool-use and agent systems extended this pattern to external application programming interfaces, model libraries, memory, and reusable behaviors \cite{yao2022react,schick2023toolformer,shen2023hugginggpt,wang2023voyager}. Skill documents are a related form of inference-time support. They give an agent procedural knowledge, local constraints, examples, and task conventions without changing model weights.

SkillsBench provides a direct benchmark for this setting. It packages expert-written skill folders with containerized tasks and deterministic verifiers across 11 domains \cite{skillsbench2026}. Its main comparison measures performance with curated skills, self-generated skills, and no skills. This article studies a narrower factor inside the same benchmark machinery: whether changing the presentation granularity of an already controlled skill document changes downstream task success.

The design question is unresolved because skill documents can be written at several levels of abstraction. A high-abstraction document can state principles and invariants. A medium-abstraction document can give procedural guidance and decision points. A low-abstraction document can provide checklist-like operational steps and recovery checks. A worked example can make a procedure more concrete, while also directing the agent toward a particular path. These are plausible design hypotheses. They require measurement under fixed tasks, fixed schedules, fixed model settings, preserved benchmark files, and inference at the task level rather than at the raw-run level.

This paper makes three contributions. It provides a controlled two-model measurement of skill presentation granularity on a frozen, oracle-validated SkillsBench subset. It reports task-level paired contrasts for two primary questions: low abstraction versus high abstraction, and medium abstraction with one worked example versus the same abstraction level without an example. It documents integrity and execution controls, including protected hashes, frozen schedules, final-row separation, and protocol amendments for transport, dependency, concurrency, and timeout incidents.

The main result separates skill availability from skill presentation granularity. Skill conditions are higher than no skill by 26.7 to 36.0 percentage points for GPT-5.5 and by 18.0 to 26.0 percentage points for DeepSeek V4-Flash. The controlled presentation contrasts are much smaller. Low abstraction differs from high abstraction by +0.7 percentage points for GPT-5.5 and -6.7 percentage points for DeepSeek V4-Flash. Adding one worked example to the medium-abstraction variant differs from the no-example variant by +0.7 and +1.3 percentage points. These estimates support a restrained conclusion. Access to task-relevant skill knowledge is associated with higher success in this subset, while the tested presentation-granularity changes have small, uncertain, and model-dependent effects.

\section{Background and Related Work}
\label{sec:background}

Prompted language models are sensitive to how task context is represented. In-context learning showed that demonstrations supplied as text can change task behavior without weight updates \cite{brown2020language}. Chain-of-thought prompting uses worked reasoning exemplars to elicit multi-step reasoning \cite{wei2022chain}. Few-shot calibration, example ordering, and prompt-format studies show that semantically close prompt variants can produce large output differences, especially when evaluation uses a single fixed prompt format \cite{zhao2021calibrate,lu2022fantastically,sclar2024quantifying}. Skill presentation belongs to this broader class of context design, but it acts on long procedural documents used by agents rather than on short task prompts alone.

LLM agent work extends prompting into interactive environments. ReAct interleaves reasoning traces and actions \cite{yao2022react}. Toolformer trains models to decide when and how to call external tools \cite{schick2023toolformer}. HuggingGPT frames an LLM as a planner over specialist models \cite{shen2023hugginggpt}. Voyager builds a reusable skill library for open-ended embodied tasks \cite{wang2023voyager}. These systems motivate evaluation settings where success depends on the model's ability to use external resources and procedural context.

Agent benchmarks now cover several operational settings. WebShop, WebArena, and Mind2Web study web interaction under natural-language goals \cite{yao2022webshop,zhou2024webarena,deng2023mind2web}. AgentBench evaluates LLM agents across multiple interactive environments \cite{liu2023agentbench}. SWE-bench focuses on real software issues with repository-level code editing \cite{jimenez2024swebench}. GAIA tests tool use, web browsing, reasoning, and multimodal assistance \cite{mialon2023gaia}. OSWorld evaluates multimodal agents in real computer environments \cite{xie2024osworld}. These benchmarks measure agent competence under task and environment variation. The present study instead holds a SkillsBench task subset fixed and manipulates the form of supplied skill knowledge.

Agent skills give reusable procedural knowledge a standard packaging form. Anthropic describes Agent Skills as organized folders of instructions, scripts, and resources that agents can discover and load dynamically \cite{anthropic2025skills}. Anthropic documentation describes Agent Skills as modular capabilities that package instructions, metadata, and optional resources, with content loaded on demand through filesystem-backed skill folders \cite{agentskills2026overview}. SkillsBench operationalises this idea with expert-written skills, deterministic task verifiers, and skill/no-skill comparisons \cite{skillsbench2026}. It reports a +16.2 percentage-point curated-skill average on its benchmark and also observes that focused skills with two to three modules outperform comprehensive documentation \cite{skillsbench2026}. Those findings motivate more targeted measurements of which properties of a skill package affect agent behavior.

The present study focuses on presentation rather than authorship. The official curated skills are preserved byte-for-byte. Four rewritten conditions are derived from compact content ledgers that represent source procedures, constraints, placeholders, and frontmatter policy. The rewritten variants change abstraction level and the presence of one worked example while keeping task identity, verifier code, model configuration, schedule, and evaluation harness fixed. This design treats a skill document as an experimental context intervention.

\section{Method}
\label{sec:method}

\subsection{Research Questions}

RQ1 estimates the task-paired effect of low-abstraction presentation relative to high-abstraction presentation. The contrast is \texttt{low\_abstract - high\_abstract}. RQ2 estimates the task-paired effect of adding one worked example to medium-abstraction guidance. The contrast is \texttt{medium\_with\_example - medium\_no\_example}. RQ3 describes whether the signs and magnitudes of RQ1 and RQ2 are consistent across the two evaluated model configurations. RQ3 is descriptive because the two model configurations are not compute-matched.

\subsection{Benchmark and Task Subset}

The experiment uses SkillsBench pinned to commit \texttt{411dc68bc00887548775c534201ca43e7e35e080}. Table~\ref{tab:design} summarizes the locked experimental design. The selected subset contains 30 tasks across 11 domains. Task selection follows a domain-balanced capped-quota rule from the locked protocol. Every selected task passed official oracle validation before the evaluation schedule was frozen. The subset is an oracle-validated measurement subset rather than an estimate of full SkillsBench performance.

The selected tasks cover cybersecurity, energy, finance, healthcare, manufacturing, mathematics, media and content, natural science, office and white collar, robotics, and software engineering. The protocol's intended difficulty mix was 14 hard and 16 medium tasks. The pinned repository metadata records 15 hard tasks and 15 medium tasks after normalizing one \texttt{middle} label to \texttt{medium}. Task identity and oracle validation are the authoritative inclusion criteria. Difficulty labels are descriptive metadata.

\begin{table}[t]
\caption{Locked experimental design}
\label{tab:design}
\centering
\small
\begin{tabularx}{\textwidth}{lX}
\toprule
\textbf{Design element} & \textbf{Locked value} \\
\midrule
Benchmark & SkillsBench at commit \texttt{411dc68bc00887548775c534201ca43e7e35e080} \\
Task subset & 30 oracle-validated tasks across 11 domains \\
Conditions & \texttt{no\_skill}, \texttt{original\_curated}, \texttt{high\_abstract}, \texttt{medium\_no\_example}, \texttt{medium\_with\_example}, \texttt{low\_abstract} \\
Models & GPT-5.5 and DeepSeek V4-Flash \\
Trials & Five trials per task-condition-model cell \\
Final rows & $30 \times 6 \times 5 \times 2 = 1{,}800$ \\
Inference unit & Task \\
Primary contrasts & RQ1 = \texttt{low\_abstract - high\_abstract}. RQ2 = \texttt{medium\_with\_example - medium\_no\_example} \\
\bottomrule
\end{tabularx}
\end{table}

\subsection{Skill Conditions and Content Control}

The six conditions separate skill availability from skill presentation. The no-skill condition removes skill instructions. The original-curated condition copies the official SkillsBench skill files byte-for-byte. The high-abstraction condition rewrites the source skill as principles, invariants, and decision rules, with no worked example. The medium-no-example condition gives procedural guidance with decision points and pitfalls, with no worked example. The medium-with-example condition uses the same medium-abstraction target while adding exactly one short worked example based on ledger-allowed content. The low-abstraction condition presents checklist-like operational guidance, concrete placeholders, step-by-step recovery checks, and no worked example.

The source tasks contain 88 official skill documents. The official curated condition preserves those files by hash. The four rewritten conditions produce 352 rewritten skill files. Each rewritten variant is generated from a compact content ledger and audited at three levels. The deterministic structural audit checks required file structure and condition constraints. The style audit checks that the requested presentation form is present and rejects obvious format violations. The ledger-coverage audit checks that essential source content is retained and that unsupported additions are not introduced. These controls reduce content drift, while leaving semantic equivalence, token length, salience, ordering, terminology, and formatting as possible sources of residual variation.

\subsection{Model Configurations and Schedule Locks}

Two model configurations are evaluated. GPT-5.5 uses medium reasoning effort. DeepSeek V4-Flash uses thinking mode with high reasoning effort through the official OpenAI-compatible Chat Completions interface. DeepSeek documentation describes thinking-mode effort controls with \texttt{high} and \texttt{max} values, and states that unsupported sampling parameters such as temperature and \texttt{top\_p} have no effect in thinking mode \cite{deepseek2026thinking}. The run used the protocol-locked high setting rather than escalating to max after the schedule was frozen. Cross-model comparisons therefore describe the evaluated configurations, not matched reasoning budgets or model-optimal settings.

The frozen design contains 30 tasks, six conditions, five trials per task-condition-model cell, and two model configurations. Each model has 900 scheduled runs, split into five official batches of 180 runs. The records specify schedule seeds of 42 for GPT-5.5 and 314 for DeepSeek V4-Flash. They do not specify an explicit API seed or a non-default GPT-5.5 sampling temperature. Repeated trials should therefore be read as repeated calls under the locked harness settings, not as seeded stochastic replicates.

\subsection{Integrity Controls and Execution Amendments}

Integrity controls were applied before, during, and after evaluation. Protected hashes covered 898 selected task files, including instructions, task metadata, tests, verifiers, oracle solutions, Dockerfiles, environment files, and official skills. All 898 protected files were present and matched their reference hashes in final validation. Generated task copies were frozen before evaluation. The Step 5 freeze recorded that only skill injection differed across condition copies. Final rows were separated from attempt rows and retry logs, so transient infrastructure attempts before verifier execution did not enter the final result files.

Several operational amendments were recorded during the campaign to preserve final-row integrity under dependency, transport, concurrency, and timeout failures. The main amendments separated Docker build acquisition from model or oracle evaluation, routed DeepSeek through its official OpenAI-compatible Chat Completions transport, allowed bounded intra-batch concurrency with single-writer result-file discipline, retried transient provider or transport failures before verifier execution, finalized official verifier timeouts as reward 0.0, and finalized explicitly operator-authorized model-side nonresponse or nontermination cases before verifier as reward 0.0. The amendments preserved task instructions, tests, verifiers, oracle solutions, Dockerfiles, thresholds, model names, reasoning settings, conditions, and schedule identity unless explicitly stated otherwise. The supplementary appendix records the detailed amendments and incident categories.

\subsection{Statistical Analysis}

The primary binary success rule is \texttt{reward == 1.0}. Fractional rewards are retained for secondary mean-reward summaries and counted as non-passes in primary pass-rate analyses. For each task-condition-model cell, the five trials are aggregated into a task-level pass rate. Condition means are the mean of 30 task-level pass rates. This aggregation follows the experimental unit. Treating all 1,800 rows as independent would overstate precision.

For each model and primary contrast, the analysis computes the mean paired task-level difference, a 95\% bootstrap confidence interval over tasks with 10,000 resamples, and a paired Monte Carlo sign-flip permutation p-value with 100,000 samples. Holm correction is applied within each model across the two primary contrasts \cite{efron1979bootstrap,good2005permutation,holm1979simple}. Reference comparisons are reported for skill conditions against no skill and rewritten variants against original curated skills. RQ3 reports cross-model sign agreement, confidence-interval overlap, and effect gaps descriptively.

A sensitivity calculation describes the precision of the realised design. With 30 paired tasks, a normal-approximation calculation using the observed task-level pass-rate contrast standard deviations gives approximate 80\% detectable differences of 8.3 to 16.0 percentage points across the four primary model-contrast combinations. The design is therefore informative for moderate contrasts, while one- or two-percentage-point effects remain below its resolving power.

\section{Results}
\label{sec:results}

\subsection{Final-Row Validation}

The final data satisfy the locked design counts. GPT-5.5 has 900 final rows, with 150 rows per condition. DeepSeek V4-Flash has 900 final rows, with 150 rows per condition. Both result files have 900 unique run identifiers, 900 unique schedule indexes, and no null rewards. GPT-5.5 has 508 success rows, 392 non-success rows, ten fractional-reward rows, and one row with \texttt{verifier\_reached=false}. DeepSeek V4-Flash has 395 success rows, 505 non-success rows, 30 fractional-reward rows, and seven rows with \texttt{verifier\_reached=false}.

\subsection{Condition-Level Outcomes}

\begin{figure}[t]
\centering
\includegraphics[width=0.95\textwidth]{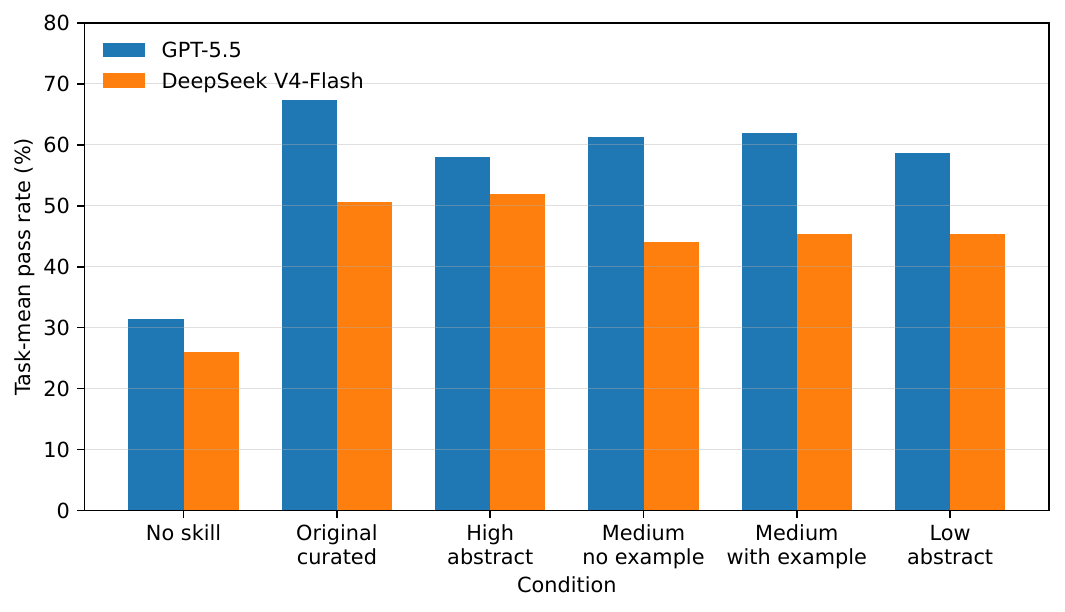}
\caption{Task-mean pass rates by model and condition}
\label{fig:condition-pass-rates}
\end{figure}

Table~\ref{tab:condition-summary} reports task-mean pass rate and secondary mean reward for each condition, and Fig.~\ref{fig:condition-pass-rates} shows the same pass rates by model and condition. The task-mean pass rate equals the descriptive run-level pass rate because every condition has 30 tasks and five trials per task.

\begin{table}[t]
\caption{Condition-level outcomes by model}
\label{tab:condition-summary}
\centering
\small
\begin{tabular}{llccc}
\toprule
\textbf{Model} & \textbf{Condition} & \textbf{Task-mean pass rate} & \textbf{Mean reward} & \textbf{Successes / trials} \\
\midrule
GPT-5.5 & \texttt{no\_skill} & 31.3\% & 0.334 & 47 / 150 \\
GPT-5.5 & \texttt{original\_curated} & 67.3\% & 0.673 & 101 / 150 \\
GPT-5.5 & \texttt{high\_abstract} & 58.0\% & 0.602 & 87 / 150 \\
GPT-5.5 & \texttt{medium\_no\_example} & 61.3\% & 0.613 & 92 / 150 \\
GPT-5.5 & \texttt{medium\_with\_example} & 62.0\% & 0.620 & 93 / 150 \\
GPT-5.5 & \texttt{low\_abstract} & 58.7\% & 0.587 & 88 / 150 \\
DeepSeek V4-Flash & \texttt{no\_skill} & 26.0\% & 0.275 & 39 / 150 \\
DeepSeek V4-Flash & \texttt{original\_curated} & 50.7\% & 0.532 & 76 / 150 \\
DeepSeek V4-Flash & \texttt{high\_abstract} & 52.0\% & 0.533 & 78 / 150 \\
DeepSeek V4-Flash & \texttt{medium\_no\_example} & 44.0\% & 0.454 & 66 / 150 \\
DeepSeek V4-Flash & \texttt{medium\_with\_example} & 45.3\% & 0.471 & 68 / 150 \\
DeepSeek V4-Flash & \texttt{low\_abstract} & 45.3\% & 0.468 & 68 / 150 \\
\bottomrule
\end{tabular}
\end{table}

\subsection{Reference Comparisons Against No Skill}

The skill-availability reference comparisons are positive for both model configurations, as reported in Table~\ref{tab:reference}. For GPT-5.5, task-level skill-versus-no-skill deltas range from +26.7 to +36.0 percentage points, and all corresponding bootstrap intervals are above zero. For DeepSeek V4-Flash, the range is +18.0 to +26.0 percentage points, and all corresponding bootstrap intervals are above zero.

\begin{table}[t]
\caption{Reference comparisons for skill conditions against no skill. Permutation $p$-values are uncorrected exploratory values and are not Holm-adjusted.}
\label{tab:reference}
\centering
\small
\begin{tabular}{llccc}
\toprule
\textbf{Model} & \textbf{Skill condition} & \textbf{Delta vs.\ no skill} & \textbf{95\% bootstrap CI} & \textbf{Permutation $p$} \\
\midrule
GPT-5.5 & \texttt{original\_curated} & +36.0 pp & [+20.7, +52.0] pp & 0.00018 \\
GPT-5.5 & \texttt{high\_abstract} & +26.7 pp & [+14.7, +40.0] pp & 0.00017 \\
GPT-5.5 & \texttt{medium\_no\_example} & +30.0 pp & [+16.7, +44.0] pp & 0.00016 \\
GPT-5.5 & \texttt{medium\_with\_example} & +30.7 pp & [+16.7, +46.0] pp & 0.00021 \\
GPT-5.5 & \texttt{low\_abstract} & +27.3 pp & [+12.7, +42.7] pp & 0.00191 \\
DeepSeek V4-Flash & \texttt{original\_curated} & +24.7 pp & [+12.7, +38.0] pp & 0.00048 \\
DeepSeek V4-Flash & \texttt{high\_abstract} & +26.0 pp & [+11.3, +41.4] pp & 0.00277 \\
DeepSeek V4-Flash & \texttt{medium\_no\_example} & +18.0 pp & [+4.7, +33.3] pp & 0.02642 \\
DeepSeek V4-Flash & \texttt{medium\_with\_example} & +19.3 pp & [+6.7, +33.3] pp & 0.01253 \\
DeepSeek V4-Flash & \texttt{low\_abstract} & +19.3 pp & [+4.7, +34.7] pp & 0.02304 \\
\bottomrule
\end{tabular}
\end{table}

The official curated skill remains a strong reference point. GPT-5.5 rewritten variants are below original curated skills by 5.3 to 9.3 percentage points. DeepSeek V4-Flash high abstraction is +1.3 percentage points above original curated skills, while the other rewritten variants are 5.3 to 6.7 percentage points below them. The rewritten-versus-original intervals cross zero in all cases.

\subsection{Primary Presentation Contrasts}

Table \ref{tab:primary} reports the two primary task-paired contrasts. RQ1 shows no stable advantage for low-abstraction guidance. GPT-5.5 is nearly unchanged at +0.7 percentage points. DeepSeek V4-Flash is lower under low abstraction than under high abstraction by -6.7 percentage points. Both bootstrap intervals include zero. The DeepSeek interval is mostly negative, but the Holm-adjusted permutation p-value is 0.2695.

\begin{table}[t]
\caption{Primary task-paired presentation contrasts}
\label{tab:primary}
\centering
\small
\begin{tabular}{llp{4.4cm}cccc}
\toprule
\textbf{Model} & \textbf{RQ} & \textbf{Contrast} & \textbf{Delta} & \textbf{95\% CI} & $p$ & \textbf{Holm $p$} \\
\midrule
GPT-5.5 & RQ1 & \texttt{low\_abstract - high\_abstract} & +0.7 pp & [-10.0, +12.0] pp & 1.0000 & 1.0000 \\
GPT-5.5 & RQ2 & \texttt{medium\_with\_example - medium\_no\_example} & +0.7 pp & [-5.3, +6.0] pp & 1.0000 & 1.0000 \\
DeepSeek V4-Flash & RQ1 & \texttt{low\_abstract - high\_abstract} & -6.7 pp & [-14.7, +0.7] pp & 0.1348 & 0.2695 \\
DeepSeek V4-Flash & RQ2 & \texttt{medium\_with\_example - medium\_no\_example} & +1.3 pp & [-6.0, +9.3] pp & 0.8715 & 0.8715 \\
\bottomrule
\end{tabular}
\end{table}

\begin{figure}[t]
\centering
\includegraphics[width=0.95\textwidth]{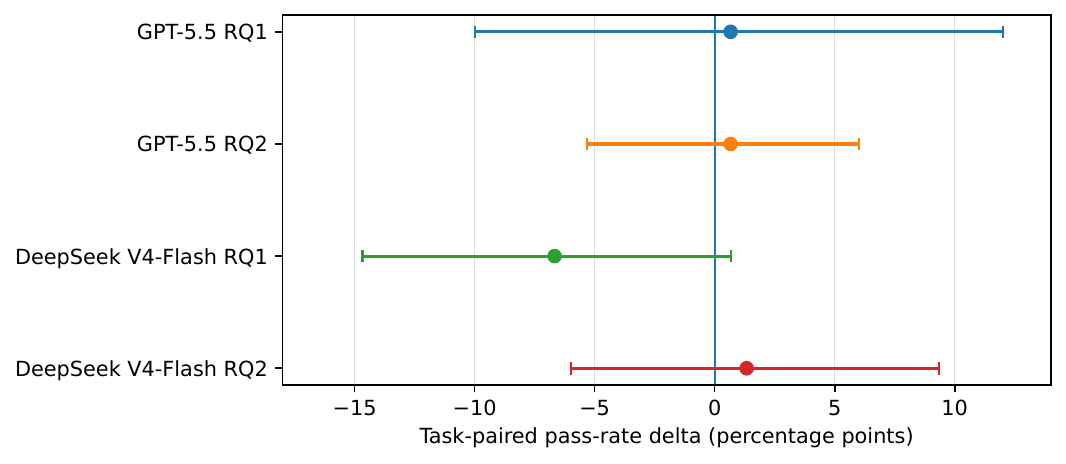}
\caption{Primary task-paired presentation contrasts with 95\% bootstrap confidence intervals}
\label{fig:primary-contrasts}
\end{figure}

RQ2 is directionally positive in both models, but the estimated effects are small. GPT-5.5 changes by +0.7 percentage points and DeepSeek V4-Flash changes by +1.3 percentage points. Both confidence intervals cross zero. The estimates are compatible with negligible worked-example effects under this protocol. Fig.~\ref{fig:primary-contrasts} shows the two primary contrasts with their 95\% bootstrap confidence intervals.

\subsection{Cross-Model Consistency and Robustness}

RQ1 changes sign across the two model configurations, as summarized in Table~\ref{tab:cross-model}. GPT-5.5 is slightly positive and DeepSeek V4-Flash is negative, with overlapping confidence intervals and an absolute effect gap of 7.3 percentage points. RQ2 has sign agreement because both model estimates are positive, but the effects are close to zero and both intervals include zero.

Mean-reward secondary contrasts preserve the substantive interpretation, as shown in Table~\ref{tab:mean-reward}. All mean-reward deltas are reported in percentage points on the 0--1 reward scale, so $-1.6$ reward points corresponds to a $-0.016$ change in mean reward. Using numeric reward rather than binary pass rate changes GPT-5.5 RQ1 from +0.7 pass-rate percentage points to $-1.6$ reward points. The sign change reflects fractional-reward rows being treated as non-passes in the binary rule and remains within the noise range of the contrast. GPT-5.5 RQ2 remains +0.7 reward points. DeepSeek RQ1 is $-6.6$ reward points, close to its binary estimate of $-6.7$ percentage points. DeepSeek RQ2 is +1.6 reward points. The secondary scores do not turn either primary contrast into a stable presentation advantage.

\begin{table}[t]
\caption{Cross-model consistency of primary contrasts}
\label{tab:cross-model}
\centering
\small
\begin{tabular}{lcccc}
\toprule
\textbf{RQ} & \textbf{GPT-5.5} & \textbf{DeepSeek} & \textbf{Sign agreement} & \textbf{CI overlap} \\
\midrule
RQ1 & +0.7 pp & -6.7 pp & No & Yes \\
RQ2 & +0.7 pp & +1.3 pp & Yes & Yes \\
\bottomrule
\end{tabular}
\end{table}

\begin{table}[t]
\caption{Secondary mean-reward contrasts for primary questions}
\label{tab:mean-reward}
\centering
\small
\begin{tabular}{llcc}
\toprule
\textbf{Model} & \textbf{RQ} & \textbf{Pass-rate delta (pp)} & \textbf{Mean-reward delta (pp, 0--1 scale)} \\
\midrule
GPT-5.5 & RQ1 & +0.7 & -1.6 \\
GPT-5.5 & RQ2 & +0.7 & +0.7 \\
DeepSeek V4-Flash & RQ1 & -6.7 & -6.6 \\
DeepSeek V4-Flash & RQ2 & +1.3 & +1.6 \\
\bottomrule
\end{tabular}
\end{table}

\subsection{Failure Annotation and Finalization Categories}

Most selected failure annotations correspond to ordinary task failures: the official verifier was reached and returned reward 0.0. The final data also include four official verifier-budget timeouts, each recorded as reward 0.0 after verifier execution. Eight final rows are explicitly operator-authorized model-side timeout or nontermination outcomes before verifier, one for GPT-5.5 and seven for DeepSeek V4-Flash. These rows are counted as model outcomes with reward 0.0, not as infrastructure retries. Transient provider, ACP, Docker, or empty pre-verifier attempts were retained in attempt and retry logs and did not produce final rows unless a numeric final outcome was obtained.

\section{Discussion}
\label{sec:discussion}

The reference comparison between skill conditions and no skill is the largest empirical signal. Both models show higher task-mean pass rates when a skill document is available. This pattern is consistent with the SkillsBench finding that task-relevant procedural knowledge can improve agent performance. The magnitude should be read as subset-specific. The present subset contains 30 oracle-validated selected tasks, while the SkillsBench headline average is computed over a broader benchmark. Model versions, harness settings, and task selection criteria also differ. The reference comparison verifies that the selected subset retains a skill-availability signal, while leaving the source of the larger magnitude unresolved.

The primary presentation contrasts are smaller and less stable than the skill-availability contrasts. Low-abstraction checklist guidance does not reliably outperform high-abstraction principle guidance. The estimate is near zero for GPT-5.5 and negative for DeepSeek V4-Flash. A possible mechanism is that detailed checklist text changes the balance between external guidance and the model's own planning. DeepSeek V4-Flash was evaluated with thinking mode at high effort, so additional low-level instructions may duplicate or constrain planning rather than improve it. The experiment did not manipulate reasoning mode or inspect internal planning traces, so this explanation remains a hypothesis.

The worked-example contrast is directionally stable but small. A single example may help by grounding an otherwise procedural document in a concrete path. The same example may also consume context or over-specify a path that does not fit the current task instance. The observed +0.7 and +1.3 percentage-point estimates are too small for a design rule. They are better treated as a candidate effect for a larger experiment that varies example number, example length, and example similarity to the target task.

The original curated skills remain an informative reference. Controlled rewrites reduce uncontrolled presentation differences, but they may remove pragmatic features of the official documents. Those features can include ordering, emphasis, naming conventions, file-local cues, and concise reminders that content ledgers do not fully capture. GPT-5.5's lower pass rates for all rewritten variants relative to original curated skills are consistent with that possibility. DeepSeek V4-Flash shows a different pattern, with high abstraction slightly above original curated skills and the other rewritten variants below them. The evidence supports caution against treating a rewritten skill as a purely equivalent presentation of the same knowledge.

Length and salience remain residual confounds in interpreting the abstraction contrast. The audits checked condition structure and ledger coverage, but abstraction level can still co-vary with document length, local emphasis, ordering, terminology, and checklist density. This limitation matters most for RQ1 because low-abstraction guidance changes both step concreteness and the amount of operational scaffolding available to the agent. The near-zero and negative RQ1 estimates support caution about a low-abstraction advantage, not a pure estimate of abstraction alone.

The execution amendments make the evaluation more auditable, while also defining operational choices that should be reproduced. Transport retries before verifier execution, bounded concurrency, Docker build separation, official verifier-timeout finalization, and operator-authorized model-side timeout finalization all affect how a long-running agent campaign becomes a clean result table. Reporting those rules avoids silent attrition. It also means follow-up studies should either reproduce the same finalization policy or report departures from it.

\section{Threats to Validity}
\label{sec:threats}

The study uses one benchmark and a 30-task subset. The subset is domain-balanced and oracle-validated, but it does not represent the full SkillsBench task distribution. The task-level sample size is 30 per model. The sensitivity calculation indicates limited resolution for effects below roughly 8 to 16 percentage points, depending on the realised contrast variance.

The content-control procedure is approximate. Content ledgers and audits reduce unsupported additions and omissions, but they cannot guarantee semantic equivalence or preserve all pragmatic cues in the original curated skills. Rewritten variants may differ in token length, salience, ordering, terminology, or formatting in ways that affect model use beyond the intended abstraction manipulation.

The two model configurations are not compute-matched. GPT-5.5 and DeepSeek V4-Flash use model-specific reasoning settings and different transport paths. DeepSeek was run with the protocol-locked high reasoning-effort setting rather than the max setting. This lock avoids post-freeze tuning, but it prevents a claim about DeepSeek max-effort behavior. Cross-model comparisons are therefore descriptive and should not be read as estimates over a population of LLM agents.

The benchmark outcome is official verifier reward, not complete real-world task correctness. Fractional rewards are treated as non-passes for the primary binary analysis, although they are retained in secondary mean-reward summaries. This choice is strict and reproducible, but it discards some gradation in task performance.

Execution amendments were necessary to complete the evaluation campaign. The amendments preserved benchmark content and final-row integrity, yet they introduce operational choices that can matter for replication. Bounded concurrency, transport retries, dependency caches, and operator-authorized model-side finalizations should be reproduced or reported explicitly in follow-up studies.

Failure annotations are qualitative and selected by model-condition strata. They help interpret incident classes and finalization rules, but they are not a complete taxonomy of all model failures.

\section{Data and Reproducibility}
\label{sec:reproducibility}

The supplementary material accompanying this article provides the aggregate analysis tables and the consolidated \texttt{analysis\_results.json}, together with the task subset, condition definitions, execution amendments, finalization categories, secondary scoring checks, and design-sensitivity calculation. The complete reproducibility package, including the two row-level JSONL result files, the per-run schedules, the protected-hash manifest, the oracle-validation report, the attempt and retry logs, the rewritten skill files and content ledgers, and the analysis and figure-generation scripts, will be released in a public repository upon publication. From these files the full analysis can be regenerated end to end. Final validation of the row-level data reports 900 GPT-5.5 rows, 900 DeepSeek V4-Flash rows, and 898 protected hashes checked. All protected files were present, all hashes matched, run identifiers and schedule indexes were unique, and every row had a numeric reward.

\section{Conclusion}
\label{sec:conclusion}

This controlled SkillsBench measurement separates skill availability from presentation granularity. Skill conditions show a positive reference signal relative to no skill for both evaluated model configurations. The primary presentation contrasts are small and uncertain. Low-abstraction guidance does not consistently outperform high-abstraction guidance, and adding one worked example to medium-abstraction guidance yields only small positive estimates. The evidence supports a restrained design implication. Task-relevant skill knowledge matters in this subset, while the tested rewrites of presentation granularity do not provide a reliable advantage over one another or over the original curated skills.

\section*{Author Biographies}

\noindent\textbf{Xiaonan Xu.} Xiaonan Xu received a graduate degree in Computer Information Technology from Northern Arizona University, Flagstaff, AZ, USA. His research interests include large language models and agent evaluation.

\noindent\textbf{Wenjing Wu.} Wenjing Wu received a graduate degree in Computer Science from the University of Colorado Boulder, Boulder, CO, USA. Her research interests include machine learning and language model applications.

\end{document}